%% file: eacl2023.tex
\pdfoutput=1

\documentclass[11pt]{article}

\usepackage{EACL2023}

\usepackage{times}
\usepackage{latexsym}
\usepackage{graphicx}
\usepackage{float}
\usepackage{makecell}
\usepackage{url}
\usepackage{hyperref}
\usepackage{amssymb}
\usepackage{amsmath}
\usepackage{booktabs} 
\usepackage{multirow}
\usepackage{dsfont}
\usepackage{xcolor}
\usepackage{ulem}

\usepackage[T1]{fontenc}

\usepackage[utf8]{inputenc}

\usepackage{microtype}
\usepackage{booktabs} 

\title{Stabilized In-Context Learning with Pre-trained \\ Language Models for Few Shot Dialogue State Tracking}

\author{Derek Chen, Kun Qian, Zhou Yu \\
  Dialogue NLP Lab \\
  Columbia University \\
  \texttt{\{dc3761, kq2157, zy2461\}@columbia.edu} \\
}

\begin{document}
\maketitle
\begin{abstract}

Prompt-based methods with large pre-trained language models (PLMs) have shown impressive unaided performance across many NLP tasks. 
These models improve even further with the addition of a few labeled in-context exemplars to guide output generation. However, for more complex tasks such as dialogue state tracking (DST), designing prompts that reliably convey the desired intent is nontrivial, leading to unstable results.
Furthermore, 
building in-context exemplars for dialogue tasks is difficult because conversational contexts are long while model input lengths are relatively short.

To overcome these issues we first adapt a meta-learning scheme to the dialogue domain which stabilizes the ability of the model to perform well under various prompts. We additionally design a novel training method to improve upon vanilla retrieval mechanisms to find ideal in-context examples. Finally, we introduce a saliency model to limit dialogue text length, allowing us to include more exemplars per query. In effect, we are able to achieve highly competitive results for few-shot DST on MultiWOZ.

\end{abstract}

\input{sections/1_intro.tex}
\input{sections/2_related.tex}
\input{sections/3_method.tex}
\input{sections/4_experiments.tex}

\input{sections/5_results.tex}
\input{sections/6_analysis.tex}

\input{sections/7_conclusion.tex}
\input{sections/8_limitation.tex}
\bibliography{anthology,custom}
\bibliographystyle{acl_natbib}
\appendix
\input{sections/appendix.tex}

\end{document}

%% file: sections/1_intro.tex
\section{Introduction}
Tremendous gains have been made on dialogue state tracking (DST) using large pre-trained language models (PLMs)~\cite{hosseini2020simple, peng2021soloist}, 
Fine-tuning such systems though require significant amounts of data, which in turn require substantial effort to collect.  Recently, prompting has emerged as a technique for achieving strong performance in a less resource intensive manner~\cite{schick-schutze-2021-exploiting, Liu2021PretrainPA}. Even better performance is possible with in-context exemplars providing a pattern for the model to follow~\cite{Brown2020gpt3}. Ideally, we should be able to apply these concepts to complex tasks like DST, but results so far have been limited~\cite{Madotto2021FewShotBP}.

\begin{figure}[ht]
  \includegraphics[width=\linewidth]{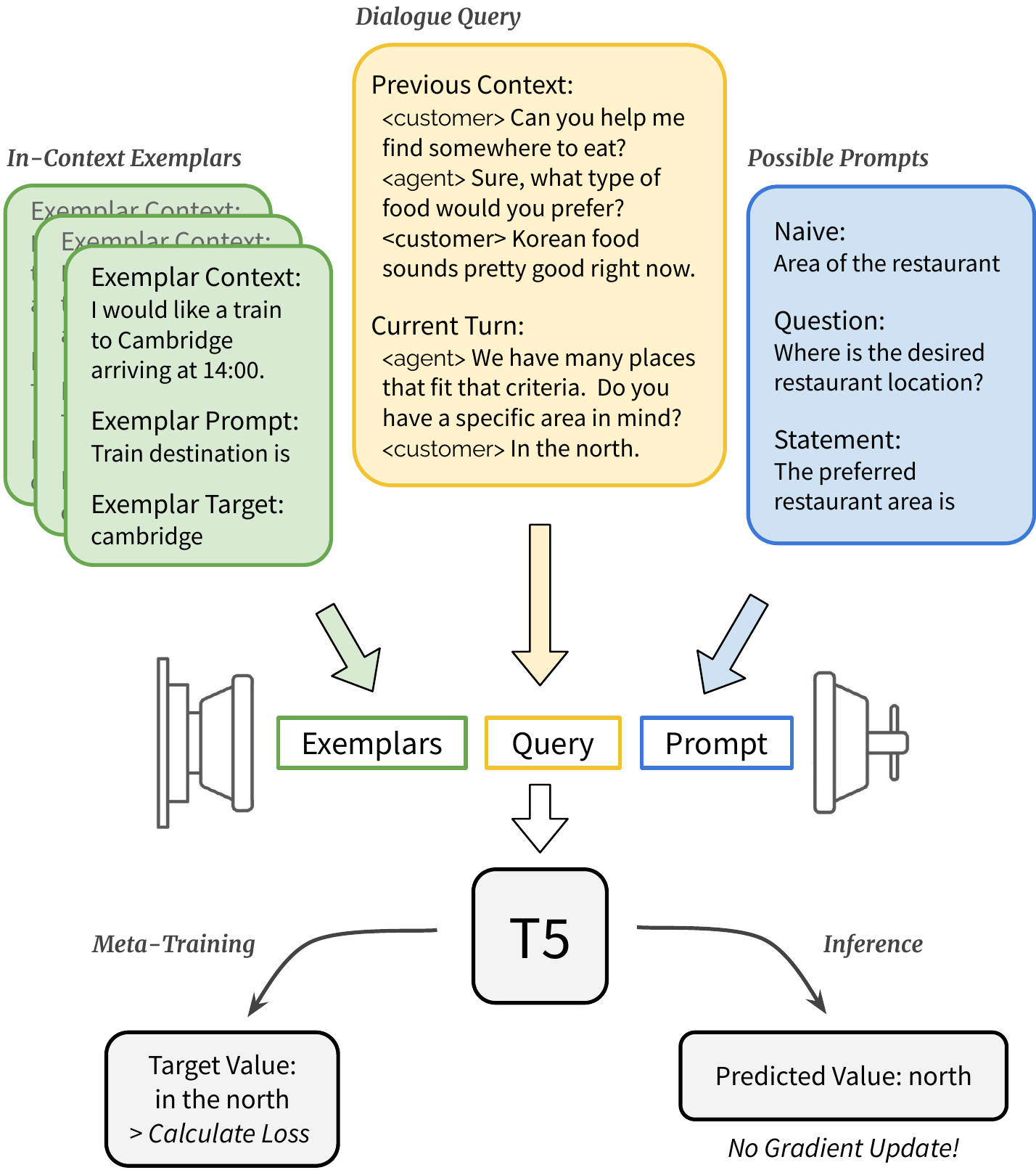}
  \caption{Our system squeezes multiple in-context exemplars, dialogue query with conversational context, and a full prompt into the finite input length of a large PLM to successfully perform few-shot dialogue state tracking, without any need for task-specific training.}
  \label{fig:front_fig}
\end{figure}

One reason for the lack of progress comes from the difficulty of hand-crafting prompts (patterns) and targets (verbalizers), which are highly sensitive to exact phrasing~\citep{lester21prompt}. 
While manually designed prompts have been found to be brittle and unstable~\cite{gu21ppt}, automatically designed prompts \cite{gao21lmbff} cannot be easily applied to DST since many slots are non-enumerable~\cite{rastogi2020sgd}.  
A second major hurdle is around dialogue sequence lengths, which are often much longer than those for other tasks ~\cite{quan2020longcontext,kottur2021dialogstitch} 
preventing the inclusion of many exemplars for guidance. Full conversations consist of long histories going back many turns, such that the context itself (sans prompt) is already capable of filling a model's entire input length. Since state tracking requires carrying over previous dialogue states, naively truncating prior context effectively equates to random guessing~\cite{heck-etal-2020-trippy, kim-etal-2020-somdst}.
A third issue is selecting the exemplars themselves. Prior work recommends choosing a representative example from each class~\cite{gao21lmbff}, but this is not possible in many cases since most domain-slot-value label combinations simply do not appear in the dataset. Moving to the few-shot scenario further exacerbates this sparsity.

Separately, recall that our main goal is to do well in \textit{few-shot} DST because we purposefully operate in a practical, low-resource data setting.  Correspondingly, we aim to achieve good results with a similar low-resource model setting where training should be possible on a single publicly-available commodity server.  This precludes the usage of gigantic models such as GPT-3, which are prohibitively expensive to train and bear high economic and environmental costs for inference alone~\cite{strubell2019energy, Bender2021energy}.

We directly tackle each of the three aforementioned issues to achieve state-of-the-art performance on MultiWOZ when restricted to models under 100 billion parameters.  To minimize prompt issues, we introduce a meta in-context learning (ICL) framework to stabilize training and reduce variance in prompt performance.  To deal with long dialogues, we are inspired by summarizaton work to condense dialogue histories and filter out non-salient sentences. Our third contribution is designing a novel loss function to train a retrieval model that selects ideal exemplars for priming our downstream model.  Our analysis and ablations show that all components help improve our state tracking performance.  Finally, we show that unlike other models which only work on specialized LMs, our proposed methods work on any sort of LM, 
and can be improved with additional training.

%% file: sections/2_related.tex
\section{Related Works}

\begin{figure*}[ht]
  \includegraphics[width=\textwidth]{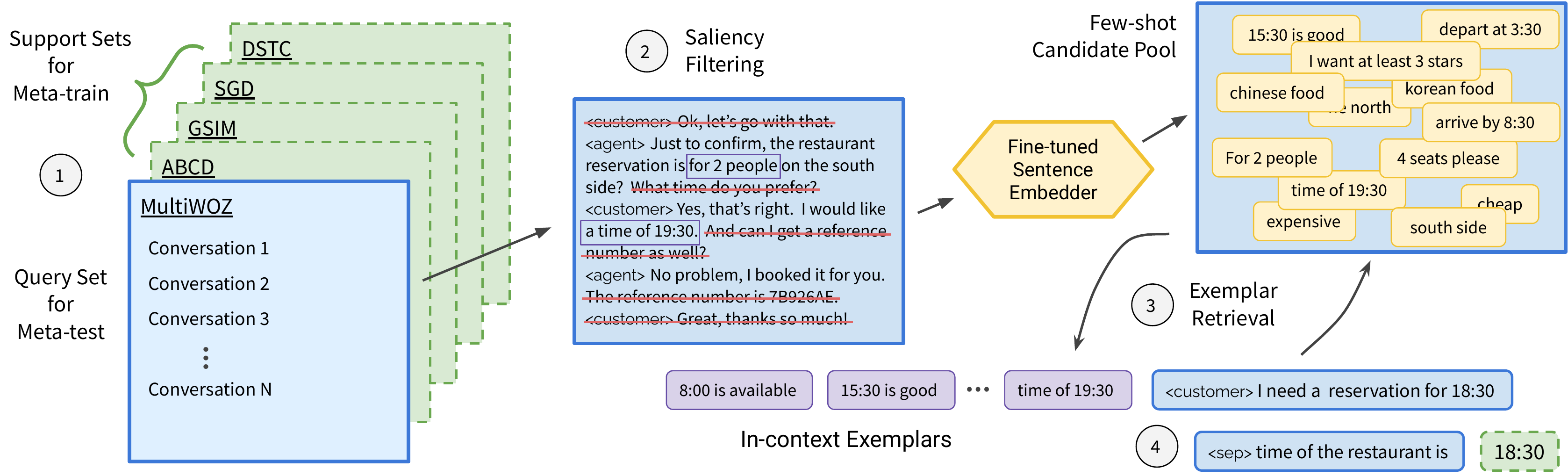}
  \caption{Our method SM2 includes (1) meta-learning with various support sets, (2) saliency filtering to remove irrelevant utterances and (3) improved exemplar retrieval from a few-shot candidate pool. Exemplars are full utterances with dialogue context, which we display as short phrases for illustrative purposes only. They are concatenated and fed into the model for prediction in Step 4. Items in green boxes, including the target value, are only available during meta-training. Purple items are raw text, while yellow ones represent their embedding vectors.}  
  \label{fig:method}
\end{figure*}

\subsection{Few-Shot Dialog State Tracking}
Nearly all recent works on dialogue state tracking leverage large pre-trained LMs to achieve good performance~\cite{heck-etal-2020-trippy, kim-etal-2020-somdst, peng2021soloist}. These methods require fine-tuning on large amounts of annotated data, whereas we hope to do well with minimal data. 

Few-shot learning can be achieved in many ways, with transfer learning probably being the most popular, where knowledge is transferred from one domain to another~\cite{wu2019trade, campagna2020transfer}.  Data augmentation also supports few-shot learning by generating additional training examples from the few-shot data~\cite{Yin2020rlaug, summerville2020plmaug, Mi21GradAug}.  Clustering techniques like prototypical networks have also shown prior success~\cite{Snell2017Prototypical}.

\subsection{Meta In-context Learning with Prompting}
This work leans on the few-shot techniques of meta-learning~\cite{Finn2017MAML} and prompting with large PLMs~\cite{Madotto2021FewShotBP}. Meta-learning allows you to get away with only a few examples at test time by pre-training a model to learn how to learn~\cite{Nichol2018OnFM}.
More recent methods which circumvent the need to calculate second-order gradients~\cite{Nichol2018Reptile} have been successfully applied to the task of DST~\cite{Dingliwal2021FewSD}, but still require fine-tuning on the query set.

Using prompts as natural language instructions have been found to work well on a wide variety of NLP tasks, including dialogue state tracking~\cite{Yang2022InversePrompt}. Prompts can be brittle though, so prompt engineering has become its own complex task with numerous ideas on finding discrete prompts~\cite{gao21lmbff} or tuning soft prompts, such as through adapters~\cite{xu2022retrieval}, prefix tuning~\cite{li2021prefix}, or prompt tuning~\cite{lester2021prompttuning}. Others have even altered the prompt structure into code in order to fit the capabilities of the network~\cite{lee2021sgpdst}. 
Inspired by the success of meta in-context learning on classification tasks~\cite{Min2021MetaICL,chen2022metaicl}, our work aims to side-step the prompt design issue altogether. Concretely, our method applies meta-learning to teach a model to recognize arbitrary instructions, thereby eliminating the need to rely on domain expertise to craft an optimal prompt.


\subsection{Exemplar Retrieval}
Lastly, our work is related to retrieval with dense vectors to find good exemplars for in-context learning~\cite{Liu2022WhatMG}. Using dense vectors for similarity search have been applied to dialogue in the past, but mainly in the context of open-domain chat~\cite{adolphs21ReasonThenRespond, komeili22internetaugmented} or knowledge-base retrieval~\cite{eric17kvret}. \citet{lee2021sgpdst} is concurrent work which leverages embeddings to search for exemplars in dialogue.

%% file: sections/3_method.tex
\section{Our Method}


This section describes our proposal of a \textbf{S}tabilized dialogue state tracker, which leverages \textbf{M}eta in-context learning, dialogue \textbf{S}ummarization and
a novel \textbf{M}ulti-part training loss for fine-tuning a retrieval model, which we refer to as \textbf{SM2} for short.

\subsection{Preliminaries}
The goal of dialogue state tracking (DST) is to extract key information from the conversation as a means of understanding the customer's intentions in each dialogue turn.  More formally, given the dialogue history $H=\{C_1, A_1, C_2, A_2, \dots, C_t\}$ composed of a series of utterances between a customer $C_i$ and an agent $A_i$, the model should predict the cumulative dialogue state up to current $t$-th turn.  This state is represented as a set of \textit{(domain, slot, value)} tuples, which our system produces by iterating over valid domain-slot pairs and then aggregating all non-null, predicted values for the given turn.  A few-shot setup only allows access to K\% of the available labeled data, with k=[1,5,10] for our experiments, where samples are randomly selected from the full labeled dataset. 
While we compare to models \textit{trained} on k-shot data, our system actually goes a step further since our eventual model receives \textit{no gradient signal} from the task-specific data and instead relies solely on in-context learning to perform inference.



\subsection{Stabilized Meta-learning}
The intuition behind prompting is that large PLMs understand instructions when written in natural language~\cite{Brown2020gpt3}. 
Thus, we write natural language patterns in an attempt to elicit the dialogue state from the model.  
However, as previously discussed, minor tweaks in prompt text may cause extreme changes in generated output, leading to highly unstable results~\cite{gu21ppt}.

Recent works on Meta-ICL~\cite{Min2021MetaICL, chen2022metaicl} have shown promise in stabilizing the variance of prompts such that crafting the perfect prompt is no longer necessary, and instead, any reasonable natural language prompt will suffice.
Classic meta-learning leverages abundant labeled data from support sets to adapt a model to quickly learn a limited-data target task, denoted as the query set.  \citet{Finn2017MAML} proposes MAML that simulates the inner adaptation step during meta-training by conducting a temporary one-step update before computing the loss.  Afterwards, a costly second-order gradient is calculated in the outer loop to train the model for faster future adaptations.  To get around the expensive loss calculation, variants such as FOMAML have since been developed~\cite{Nichol2018OnFM, Nichol2018Reptile}. Meta-ICL ingeniously avoids this calculation by replacing the inner adaptation step with in-context learning, which does not require computing gradients! More specifically, in-context learning refers to the use of exemplars to guide the model towards exhibiting ideal behavior. Critically, these exemplars are included as part of the standard model input and thus do not require gradient updates to provide a useful boost.

Following the idea of Meta-ICL, we consider each dataset as a single task and treat MultiWOZ as the held out target task. 
Specifically, all support datasets are transformed into the DST format for meta-training, where the in-context inner loop consists of support set training examples. 
Although the model does not learn about the query set in meta-training, it \textit{is} familiarizing itself with complex DST prompts during that time, allowing it to quickly adapt to the target task in meta-testing.
Furthermore, since the prompt meaning is learned during meta-training, theoretically any prompt can be used to instruct the model, including prompts constructed from random tokens (See Table~\ref{table:prompt}).

\subsection{Dialogue Compression}
Condensing the dialogue context not only fits more exemplars into the model input sequence, but also helps the model focus on more relevant text for predicting dialogue states. We introduce two general ideas under the umbrella of compressing long dialogues into shorter input sequences.

\paragraph{Context Summarization}
As the task name implies, DST requires tracking dialogue states over long periods of time, including slot-values that were carried over from the start of the conversation.  Indeed, initial experiments validated a monotonic decrease in joint goal accuracy as each marginal utterance was removed.  Therefore, as an alternative to simply removing prior utterances, we propose summarizing the dialogue history instead.  The summary of all prior turns is represented as the predicted dialogue state up to that point, which is represented as a series of (domain, slot, value) tuples.  We tried further limiting the input length by only including state tuples directly related to the current slot prediction, but surprisingly found that this formulation of the summary fared worse.

\paragraph{Saliency Filtering}
Many sentences within a conversation do not contain valuable information, such as "Thanks, that is all I need today." or "Good bye".  In order to filter away these lines, the first instinct is to train a large model, but our situation only has access to a few labeled examples, so to keep things simple, we instead gather a small handful of  heuristics to identify non-salient utterances.  For example, lines that discuss a "reference number" or are excessively terse are targeted for removal.
We verify the performance of our heuristics on the limited few-shot examples, where we heavily weight the model's recall of salient utterances over its precision. 
We take a very conservative approach since accidentally dropping a single relevant sentence can cause a severe penalty in joint goal accuracy. 

\subsection{Multi-part Retrieval Training}
\label{sec:embedder}
Exemplars are the only guiding signal when dealing with in-context learning, so selecting quality cases is of utmost importance.
To do so, we 
fine-tune the sentence embedder used during retrieval by taking advantage of the limited, few-shot data available. 

\paragraph{Exemplar Retrieval}
Exemplars are retrieved based on their proximity to the query example. Concretely, we first encode all available exemplars into a shared embedding space using a SBERT embedder~\cite{reimers2019sbert} where the raw text fed into the embedder is the exemplar's dialogue history.  For each incoming query, we encode the instance in the same manner, and then compare their embeddings to rank the closest exemplars in the few-shot candidate pool (Step 3 in Figure~\ref{fig:method}).
Finally, we keep pulling exemplars from the top of the stack to feed into the model until the entire context length of 512 is at capacity.  Since the exemplar embeddings are pre-computed, looking for similar exemplars during inference is a very quick operation. 

\paragraph{Embedder Fine-tuning}
To improve the performance of our retrieval model, we explore two categories of training techniques.  
Inspired by the rise of contrastive learning~\cite{Hadsell2006contrastive} as a pre-training method for NLP tasks~\cite{gao21simcse, karpukhin20dpr}, we first study a \textsc{Contrastive} loss which brings positive examples closer together while pushing negative examples further apart.  In our case, 
exemplars sharing the same domain and slot are positive (Y=0) while all others are negative (Y=1). The loss becomes:

\begin{align*}
    \textrm{Loss}(i,j)=\ \  & \frac{1-Y}{2}[dist(z_i,z_j)]^2\ \ + \\
    & \ \frac{Y}{2}\{max(0,m-dist(z_i,z_j)))\}^2
\end{align*}

where $z_i$ represents the embedding vector for utterance $i$ while $m$ is a margin, set to 1. We explored various distance functions (e.g. euclidean) and found that distance based on cosine similarity worked best: 
\begin{align*}
dist(z_i\cdot z_j)=1-\frac{z_i\cdot z_j}{|z_i|\cdot|z_j|}
\end{align*}
Since we retrieve exemplars based on cosine score, we can directly optimize for this as second technique with a \textsc{Mean-Squared Error} loss.
More specifically, the positive pair is assigned a target score of 1 when the two examples share the same domain and slot and 0 otherwise, mirroring the setup of the contrastive loss. The model's predicted cosine score is then compared against this target to calculate an averaged L2-loss.  We generate $\kappa$ pairs for each of $N$ exemplars, and train our ranker with:
\begin{align*}
L(i,j) =\frac{1}{NK}\sum_{i=1}^N\sum_{j=1}^K ||\textrm{Target}(i,j)-\textrm{Pred}(i,j)||^2
\end{align*}

\paragraph{Multi-part Modification}
The standard method for selecting negatives has a few drawbacks since all negatives are treated the same.  While this is necessary for unsupervised contrastive learning, our case deals with labeled exemplars.  Even binary labels would provide a useful training signal, but we even have varying degrees of similarity.  In particular, a positive example would be an exemplar that has a matching domain, slot and value.  However, exemplars that contain a matching domain or slot still deserves partial consideration rather than being deemed a pure negative example. Consequently, we introduce a \textsc{Multi-Contrastive} loss where the different elements of domain, slot and value are considered positive attributes, weighted with their respective lambdas.  These coefficients were chosen by tuning on a held-out development set:
\begin{multline*}
   \textrm{Loss}(i,j)=  \frac{\lambda_d+\lambda_s+\lambda_v}{4}[dist(z_i,z_j)]^2+\\
   \frac{\lambda_n}{4}\{max(0,m-dist(z_i,z_j)))\}^2
\end{multline*}
where:
\begin{multline*}
\lambda_d=3,\quad \lambda_s=7,\quad \lambda_v=10 \\
\lambda_n=1.0,\quad margin=1.0
\end{multline*}
For a final loss function, we also test a novel cosine similarity loss where the target label is modified to include multiple parts, \textsc{Multi-MSE}.  The target is altered such that a matching domain for each pair gets $\lambda_d=0.3$, a matching slot receives another $\lambda_s=0.3$ boost and matching values get an additional $\lambda_v=0.4$, where the weights are derived by tuning on the dev set.  The final target score is the cumulative sum of the three components - positive pairs sharing all elements get a full score of 1, negative pairs with no matching elements receive a 0, and most pairs lie somewhere in the middle.
\begin{align*}
\text{Target}(i, j) =  \sum_e \lambda_e [\mathds{1}\{ e_i = e_j\}], \forall e \in \{d,s,v\} \\
\text{s.t.} \qquad \lambda_d + \lambda_s + \lambda_v = 1
\end{align*}

\subsection{Model Input}
\label{sec:input}
The eventual sequence we feed into the model takes all of the above ideas into account.  
We start with a context summary represented as the predicted dialogue state, followed by the current turn which consists of two utterances.  Each utterance includes a special $\texttt{\small<agent>}$ or $\texttt{\small<customer>}$ token for the respective speaker. Next, a separator token is added, along with a discrete prompt describing the domain and slot.  Lastly, we prepend as many exemplars as we can fit into the model maximum token length, truncating from the beginning when necessary. This results in a final model input of:
\begin{equation*}\begin{split}\begin{gathered} \relax
[N \ exemplars][prev\_dialog\_state][agent\_utt] \\
[customer\_utt]<sep>[prompt][value]
\end{gathered}\end{split}\end{equation*}
Notably, the final $\texttt{\small[value]}$ token is only present during meta-training, and belongs to the support datasets.  This value is precisely what we hope to predict when testing the left out query set.

%% file: sections/4_experiments.tex
\section{Experiments}
This section outlines our training implementation details as well as key experiments.

\begin{table}[t]
\small
\centering
\begin{tabular}{l|c|c|c}
\toprule
Dataset & \# Dialogs & \# Domains & \# Slots \\\midrule
MultiWOZ & 8,438       & 7        & 24     \\
SGD      & 16,142      & 16       & 214    \\
GSIM     & 1,500       & 2        & 13     \\
DSTC2    & 1,612       & 1        & 8      \\
ABCD     & 8,034       & 30       & 231    \\
\bottomrule
\end{tabular}
\caption{Statistics of involved task-oriented dialogue datasets. Note that the numbers reported
are for the training portions for all datasets.}
\label{tab:datasets}
\end{table}

\subsection{Training Setup}
We consider Schema Guided Dialogue (SGD)~\cite{rastogi2020sgd}, DSTC2 ~\cite{henderson2014dstc2}, Action-Based Conversations Dataset (ABCD)~\cite{Chen2021ABCD}, and Google Simulated Chat (GSIM)~\cite{shah2018gsim} as support sets (listed in Table~\ref{tab:datasets}).  We then use MultiWOZ 2.1 ~\cite{budzianowski18multiwoz, eric2019multiwoz} as a query set, as well as MultiWOZ 2.4 ~\citep{zang20mwoz22} which is the cleanest version of MultiWOZ at time of writing.  All datasets have dialogue compression techniques applied and use the best performing embedder for exemplar retrieval.   

For our training we use T5~\cite{2020t5} with both the three and eleven billion parameters versions (T5-3b/T5-11b), where our best models are selected through early stopping on validation data.  We set the learning rate as $3e-4$, employ an Adafactor~\cite{Shazeer2018AdafactorAL} optimizer and cosine scheduler with warmup of 10,000 steps.  Our best system uses an ensemble of exemplar embedders that were trained with of $\kappa=[20,30,40]$ and learning rate of $3e-5$. More details can be found in Appendix~\ref{sec:hyperparameters}. 

\subsection{Prompt Variations}
Model training can be considered stable if different prompts produce similar outcomes. To test this, we collect six prompts based on common sense and prior work.  As much as possible, we use prompts designed by others to avoid biasing the rankings. 

Since LMs supposedly operate on prompts as continuation of natural language, the (a) \textit{Statement} prompt takes the form `The restaurant cuisine is <blank>', where we hope the model completes the sentence with the correct slot-value.  (b) A \textit{Question} prompt reverses the meaning with `What is the restaurant cuisine?'  (c) \textit{Schema} comes from \cite{lee2021sgpdst} and MWOZ 2.2 descriptions, which aims to provide the model with the maximum amount of information.  It includes a special token, name, and full description for both the domain and slot. (See Table~\ref{table:prompt}) (d) \textit{Naive} takes the opposite approach by simply following the format of ``<slot> of the <domain> is <blank>''.  (e) Taken even further, the \textit{None} prompt does not use any natural language at all, instead opting to only include the domain and slot name for evaluation purposes. (f) Finally, we include a \textit{Random} prompt which drops any notion of semantics by replacing the domain with a random color and the slot with a random animal.  To empathize with the difficulty of hand-engineering a prompt, note that each option (except for random) seems reasonable, and it is hard to know a priori which one works best. 

\begin{table}[t]
\small
\centering
\begin{tabular}{l|p{4.9cm}}
\toprule
Prompt Style & Prompt Example \\
\midrule
Statement    &   ``The destined location of the taxi is''            \\
Question     &   ``Where is the destination of the taxi ?''           \\
Schema       &  ``<domain> taxi - rent cheap cabs to avoid traffic <slot> destination - what place you want the taxi to take you''            \\
Naive        &  ``destination of the taxi is''              \\
None         &   ``taxi destination''      \\
Random       &   ``blue cobra'' \\
\bottomrule
\end{tabular}
\caption{Examples for different prompt styles. Here we consider a domain of ``taxi'' and a slot of ``destination''.}
\label{table:prompt}
\end{table}
As a baseline, we start with in-context learning without meta-training.  We feed in the prompts directly and measure their variance as the standard deviation among scores. Then, we perform meta-learning with all prompts again and measure their results, where we expect that the variance among the scores has now decreased. 


\begin{table}[t]
\small
\centering
\begin{tabular}{l|c|c|c}
\toprule
                     & MRR@10      & NDCG@10      & MAP@100 \\
\midrule
Default              & 16.7\%     & 9.59\%      & 1.81\% \\
Contrastive          & 17.4\%     & 10.6\%      & 2.28\% \\
Multi-contrast       & 17.1\%     & 9.89\%      & 1.90\% \\
Mean Squared         & 25.1\%     & 15.5\%      & 3.31\% \\
Multi-MSE    & \textbf{26.8\% }    & \textbf{18.4\%}      & \textbf{5.24\% }\\
\bottomrule
\end{tabular}
\caption{Results of fine-tuning the sentence embedder with various loss functions. Multi-part cosine is best.}
\label{table:retrieval}
\end{table}

\subsection{Filtering Threshold}
In order to verify that our saliency model successfully removes irrelevant sentences, we employ two experts to annotate 50 dialogs, which is well below the allowed 1\% of few-shot data. 
We then run the saliency model on this tiny evaluation set with different filtering thresholds, ranging from 0.1 to 0.9, with results illustrated in Figure~\ref{fig:filter}. As the threshold increases, only sentences with high relevance are left, as evidenced by high precision and low recall. A maximum F1-score is reached at 0.6, but we would rather keep all relevant sentences at the expense of amassing a handful of irrelevant sentences than to risk missing important information. As a result, we choose 0.4 as the filtering threshold, which achieves a recall of 0.998 and acceptably high precision. Qualitative examples of irrelevant sentences that were removed can be found in section~\ref{sec:case study}.

\begin{table*}[t]
\small
\centering
\begin{tabular}{l|c|c|c|c|c|c|c}
\toprule
\multirow{2}{*}{Models}&Parameter&\multicolumn{3}{c|}{MultiWOZ2.1}&\multicolumn{3}{c}{MultiWOZ2.4}\\
\cmidrule{3-8}
   & Size & 1\% & 5\% & 10\% & 1\% & 5\% & 10\% \\
\midrule
TRADE~\cite{wu2019trade}   &\multirow{4}{*}{<1B}  &12.58   &31.17   & 36.18 & -&- &-  \\
SGPDST~\cite{lee2021sgpdst} &    &32.11   & 43.14  & \textbf{46.92} & -&- &-  \\
DS2-BART~\cite{shin2022ds2} & & 28.25  &37.71 & 40.29  & 30.55 & 42.53 & 41.73 \\
DS2-T5~\cite{shin2022ds2}&  & 33.76 & 44.20   & 45.38  & 36.76 & 49.89 & 51.05 \\
\midrule
IC-DST GPT-Neo 2.7b~\cite{Hu2022InContextLF}  &\multirow{6}{*}{<100B} &  16.70  & 26.90   & 31.65  & 17.36 & 29.62 & 34.38 \\
IC-DST CodeGen 2.7b~\cite{Hu2022InContextLF}  & & 20.72   & 29.62   & 33.81  & 21.87 & 33.16 & 37.45 \\
SM2-3b (Our Method) & & 38.06   & 39.94   & 39.85  & 37.59 & 49.22 & 50.33\\
\ \     - Saliency Filtering       & & 36.11   & 38.26   & 38.63  & - & - & - \\
\ \     - Context Summarization    & & 37.02   & 37.83   & 37.80  & - & - & - \\
\ \     - Embedder Fine-tuning     & & 27.15   & 30.88   & 31.40  & - & - & - \\
SM2-11b (Our Method)  & & \textbf{38.36}   & \textbf{44.64}   & 46.02  & \textbf{40.03} & \textbf{51.14} & \textbf{51.97} \\
\midrule
IC-DST Codex-davinc 175b~\cite{Hu2022InContextLF} &>100B & 43.13   & 47.08   & 48.67  & 48.35 & 55.43 & 56.88 \\
\bottomrule
\end{tabular}
\caption{DST performance using 1\%, 5\% and 10\% of the training set. Naive prompt used for our method. Bolded numbers indicate highest performance on models under 100 billion parameters.  Note that models <1B params fine-tune on task data. Ablation results are also included for dialogue compression and embedder training.}
\label{table:dst}
\end{table*}

\subsection{Retrieval Methods}
We adapt SBERT~\cite{reimers2019sbert} to our DST task with four different objective functions: standard contrastive loss, multi-part contrastive loss, binary cosine similarity loss and multi-part cosine similarity loss. We test with number of pairs per exemplar in a range from 10 to 100 in increments of ten.  We found $\kappa=30$ to work best, which we use moving forward.  As a control, we also include the default SBERT model without any further fine-tuning.  We evaluate the results of training on the few-shot examples with Mean Recipricol Rank (MRR@10), Normalized Discounted Cumulative Gain (NDCG@10) and Maximum Average Precision (MAP@100) as our metrics.

As is shown in Table~\ref{table:retrieval}, the multi-part cosine loss showcases the strongest ability to select meaningful exemplars.  This shows the benefit of providing partial credit to all elements of the dialogue state.  Surprisingly though, the multi-part contrastive loss underperformed. Preliminary error analysis revealed negative examples were successfully separated from positive examples, but the different positive examples were mixed together.
We adopt the embedder trained with the \textsc{Multi-MSE} for all remaining experiments.

\begin{table*}[t]
\footnotesize
\centering
\begin{tabular}{l|cccccc|c}
\toprule
Prompt Style   & None & Naive & Schema & Statement & Question & Random & STDEV \\
\midrule
Fine-Tune      & 35.3  & 39.2   & 38.7  & 41.1     & 39.3    & 24.7   & 6.02     \\
In-Context     & 17.5  & 19.9   & 14.6  & 18.9     & 12.4    & 4.80   & 5.58     \\
Pre-train      & 31.8  & 35.4   & 28.2  & 27.8     & 34.6    & 17.2   & 6.65     \\
SM2 T5-3b      & 33.9  & 39.9   & 30.0  & 38.2     & 35.6    & 33.1   & 3.58     \\
SM2 GPT-XL     & 9.70  & 8.70   & 8.50  & 11.4     & 8.90    & 1.20   & 3.53     \\
\bottomrule
\end{tabular}
\caption{Joint goal accuracy over different prompt styles. Models trained with 5\% of training data. The backbone model of Fine-tune and In-Context is T5-3b. Instability is measured as standard deviation of the accuracy scores.}
\label{table:stability}
\end{table*}

%% file: sections/5_results.tex
\section{Results and Analysis}
The goal of this work is to achieve strong results on DST without worrying about tedious prompt-engineering.  Consequently, we first analyze the ability of the best performing models and then discuss performance stability across different prompts.

\subsection{Main Results}
Table~\ref{table:dst} shows that methods based on in-context learning clearly surpass those based on fine-tuning with few-shot data, as evidenced by the strong performance of SM2 as well as the concurrent work of IC-DST~\cite{Hu2022InContextLF}. In fact, our SM2-11b model is able to achieve the best joint goal accuracy on MultiWOZ 2.1 and 2.4 for most few-shot splits, when focused on models less than 100B parameters.  Furthermore, when considering just models operating with in-context learning, SM2-3b greatly outperforms the IC-DST 2.7b models in the same order of magnitude. 
We note that our method is agnostic to model size, so it is certainly possible to combine them with systems larger than 100B params.  Doing so would likely yield strong performance without sacrificing stability.

On that note, Table~\ref{table:stability} shows that models trained with SM2 exhibit roughly a 2x reduction in variance over models trained under other regimes.  While fine-tuning on certain prompts produces some of the highest scores we observe, other prompts yield some of the lowest, highlighting how hand-crafting prompts are wrought with danger. 
The instability is most pronounced for the random prompt, which meta-learning is able to smooth over. 
Also worth noting is that meta-learning from SM2 is able to stabilize prompt performance across multiple model types, including sequence-to-sequence (row 4) or auto-regressive LMs (row 5).  This is in contrast to purely in-context models, such as those which were pre-trained on code and must always obey a rigid coding structure during inference.

\subsection{Ablation Study}
To evaluate the different contributions, we run three ablation experiments, each of which removes one of the key components of SM2.  The results presented in Table~\ref{table:dst} show that each change makes a noticeable impact. 
Without saliency filtering, model performance drops by a small, but consistent amount of roughly 1-2\%.  
Disabling context summarization means truncating dialogue history to four utterances and precluding previous dialogue state, which causes an even bigger decrease in accuracy.  
Using the default SBERT embedder deals the most damage of all, leading to a nearly 10\% drop. This suggests that exemplar selection is most critical for in-context learning methods.

The proposed ideas are also independently applicable to other NLP tasks. For example, compressing inputs to fit more exemplars into an model input sequence can be applied to dialogue generation with large LMs or even reading compression, which requires reasoning over long supporting paragraphs.  A multi-part training mechanism can be applied to tasks that contain multiple elements, such as the premise, hypothesis and labels of NLI.

\begin{table*}[ht]
\centering
\footnotesize
\begin{tabular}{l|p{5cm}|p{5cm}|p{1cm}}
\toprule
\multicolumn{4}{c}{Exemplar Retrieval}\\
\midrule
Dialog ID & Target Utterance & Exemplar & Score \\
\midrule
\multirow{7}{*}{SSNG0074.json}
&\multirow{7}{5cm}{I am looking for a restaurant in the \textcolor{blue}{\textbf{moderate price range}} that serves \textcolor{red}{\textbf{bistro type food}}.} 
&E1: I would love to help. any particular food you'd like? \textcolor{red}{\textbf{no}}, I'd just like for it to be in the east and \textcolor{blue}{\textbf{moderately priced}}.
&0.738\\
\cmidrule{3-4}
&&E2: Seventeen locations meet your criteria. Would you prefer a guesthouse or a hotel? A hotel is fine whichever you recommend.
& -0.074\\
\midrule
\multicolumn{4}{c}{Saliency Filtering}\\
\midrule
\multirow{3}{*}{PMUL0287.json}
&\multicolumn{3}{c}{\multirow{3}{11cm}{<Agent>: The phone number is 01223259988. <User>: \sout{Perfect.} Can you help me with a reservation for 6 people at 14:30 this coming sunday? \sout{And please make sure I have a confirmation number to use.} \sout{<Agent>:our reservation is set!}}}\\
\\
\\
\midrule
\multirow{4}{*}{PMUL1635.json}
&\multicolumn{3}{c}{\multirow{4}{11cm}{<Agent>: What day will you be staying? <User>: Friday and Can you book it for me and get a reference number ?\sout{ <Agent>:Booking was successful.} \sout{Reference number is : BMUKPTG6.} \sout{Can I help you with anything else today?} <User>: I am looking to book a train that is leaving from Cambridge to Bishops Stortford on Friday.}}\\
\\
\\
\\
\bottomrule
\end{tabular}
\caption{Examples of how exemplar retrieval and saliency filtering operate. Same colored text represents matching domain and slots. The strikethrough of text means removal of the irrelevant sentence by the saliency model.}
\label{table:qualitative}
\end{table*}

%% file: sections/6_analysis.tex
\subsection{Additional Discussion}  

We now turn our attention to the impact of different training regimes, as shown in Table~\ref{table:stability}.  Fine-tuning (row 1) serves as an oracle since it represents training directly on the data in the target domain. Unsurprisingly, SM2 reaches lower average results in comparison.  In contrast, SM2 significantly outperforms in-context learning (row 2) since neither perform gradient updates, while SM2 includes a meta-learning stage. Finally, to disentangle the effects of pre-training and meta-ICL, we also compare against a baseline which does not perform in-context learning (row 3). 
Rather than learning the prompts, this baseline instead simply performs transfer learning from the source datasets to the target dataset.  Such a setup does not work as well due to the domain shift from the source distribution to the target distribution. 

Digging deeper, we notice that our method displays a meaningful jump in performance when going from 1\% to 5\% data, but not much when going to 10\%.  The increased amount of data fails to provide much marginal value since the exemplars being selected did not change much despite choosing from a larger candidate pool.  Instead, the finite sequence length became the bottleneck on downstream accuracy. 

The performance of the in-context methods are interesting in their own right. Statement prompt does best, while Random does worst, but despite having no training, is well above chance. 
This surprising result confirms other research on prompt analysis, which found that large PLMs sometimes perform \textit{too well}, implying that the models are actually paying attention to superficial cues rather than truly understanding the text within a prompt~\cite{webson22understand, kavumba2022clueless}.

\subsection{Qualitative Analysis}
\label{sec:case study}

The top half of Table~\ref{table:qualitative} shows an utterance with ``\textit{domain=restaurant}'' and ``\textit{slots=price range, food type}''. 
Despite having minimal n-gram overlap with the example, the first exemplar E1 receives a high score by matching the same domain and slot of the target utterance. On the other hand, the second exemplar E2 discusses an entirely different topic, producing a low score.  This demonstrates the effectiveness of the sentence embedder in distinguishing the value of these exemplars.
The bottom half of Table~\ref{table:qualitative} shows how the saliency model successfully conserves a large amount of token space.
Short sentences and those void of any dialog state information are safe for removal.  When all sentences in an utterance are filtered, then we also remove the associated speaker token. Despite our conservative thresholds, the majority of useless information is successfully trimmed out to allow the model to focus on the most pertinent areas instead. 

%% file: sections/7_conclusion.tex
\section{Conclusion}
In this paper, we presented a method of performing few-shot dialogue state tracking by leveraging large pre-trained LMs with prompts.  Our technique does not require any gradient-based training for the target task and instead relies on in-context learning to guide model generation. 
To enable success in this low-resource setting, we stabilize 
training across prompts with Meta-ICL, apply saliency filtering and context summarization to reduce dialogue length, and fine-tune a sentence embedder with a custom loss objective to improve exemplar retrieval.
These techniques combined allow us to reach state-of-the-art results on MultiWOZ when limited to models under 100 billion parameters.

Moving forward, we plan to explore techniques that push model and data efficiency even further.  Distillation and pruning can lead to much fewer model parameters, while
numerous data augmentation techniques seem promising in maximizing the advantage of limited labeled data.
Lastly, rather than meta-learning across different dialog domains, we also would like to explore meta-train model with different prompt styles. With the current framework, the prompt used in inference is required to be the same as the training. However, we might want to use flexible prompts in practice. Consequently, we could meta-train across different prompt styles to allow the model to quickly learn a new prompt style during inference.

%% file: sections/8_limitation.tex
\section{Limitations}
Our method is model-agnostic and can be combined with larger pre-trained model over 100 billion parameters for further improvement on DST task. However, due the budget limit, this is unlikely to be directly validated. Ironically, our method also has the limitation that it cannot be combined with smaller models since the emergent behavior of being to understand prompts only seems to occur with sufficiently large pre-trained models.

Separately, the proposed saliency filtering and the exemplar retrieval module are designed based on the dialog state tracking task, but not specifically for the MultiWOZ dataset. As a result, we planned to apply our framework to other task-oriented dialog datasets, e.g. SGD~\cite{rastogi2020sgd} to verify that our framework is generalizable, but have not done so yet due to time constraints.  We also ran our experiments with a different model type in GPT-XL, but did not have a chance to properly tune the parameters, leading to low performance.  

We would have liked to run our experiments with different random seeds.  Considering the stability of our framework among different prompt styles, different random seeds should not cause high variance. However, we still need to run experiments to verify this assumption.  

\newpage

%% file: sections/appendix.tex
\newpage

\section{Loss Functions}
\label{sec:appendix}
\citet{gao21simcse} proposes a softmax-based contrastive loss:
$$L_i=-log\frac{e^{sim(\textbf{h}_i,\textbf{h}_i^+)/\tau}}{\sum_{j=1}^{N} e^{sim(\textbf{h}_i,\textbf{h}_j^{+})/\tau}}$$
which is popular among NLP tasks. However, this loss function requires extremely large batch sizes to work well~\cite{chen20simclr}. This is especially difficult for us since we specifically target a low-resource setting with small GPU memory requirements. More critically, this softmax contrastive loss views all negatives as being the same. However, in the case of dialog state tracking, where dialog state is represented as (domain, slot, value), the matching is decided at three levels. For example, two dialogue examples can (and should) be considered a negative pair when they have different values for all three elements. In another case though, they might be considered a negative pair by not having matching ``value'', but still sharing the same ``domain'' and ``slot''. The softmax constrastive loss considers these two cases as the same, which is not ideal for the DST task.  Therefore, we implement the for our experiments.  The classic max-margin contrastive loss~\cite{Hadsell2006contrastive} is also unable to make a clear distinction for partial credit either, but should be able to when the loss is the sum of multiple elements.  Therefore, we use the max-margin loss for our experiments.

\section{Filtering Results}
\begin{figure}[ht]
  \includegraphics[width=\columnwidth]{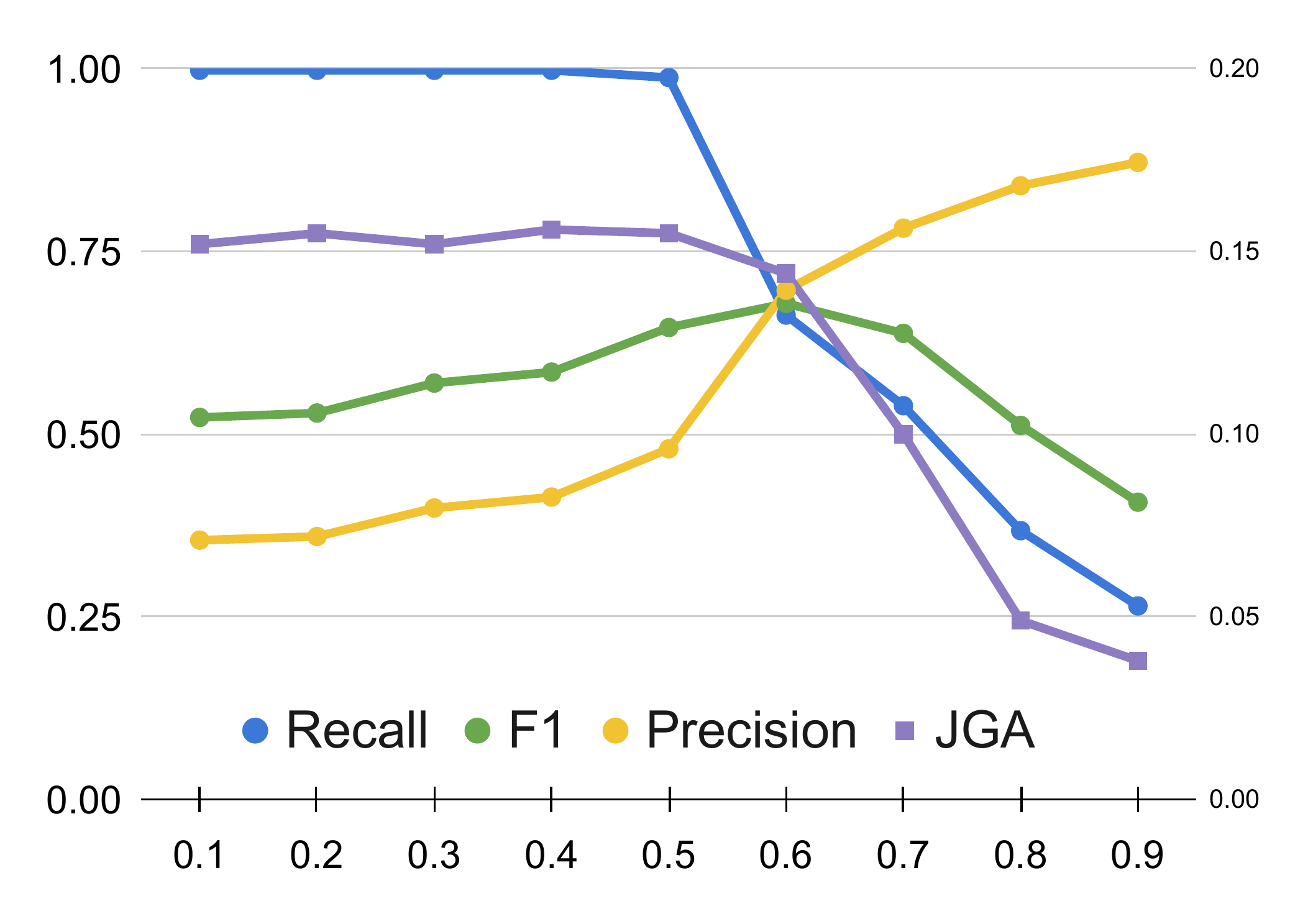}
  \caption{Graph of precision, recall and F1 when varying the acceptance threshold. Joint goal accuracy (JGA) correlates closely with recall due to the nature of DST.}
  \label{fig:filter}
\end{figure}

\section{Other Implementation Details}
\label{sec:hyperparameters}
In this section, we introduce more implementation details. For training, we search the learning rate within the interval [3e-5, 1e-4, 3e-4, 1e-3, 3e-3].  In order to deploy large pre-trained models like T5-3b and T5-11b, we first adjust the batch size. To achieve a balance between GPU memory consumption and batch performance, we alter the number of gradient accumulation steps to maintain a consistent effective batch size of 64 across runs. Furthermore, we also change everything into bitfloat 16 (BF16) and adopt AdaFactor as the optimizer to lower the number of parameters.

We additionally perform ensemble decoding for multiple times using different retrieval embedders.  These sentence embedders are distinguished by being trained on different levels of kappa, where we end up choosing embedders trained with kappa of [20,30,40].  These values were selected since they were the models which had the best results as measured by MRR@10 and MAP@10.  We run exemplar retrieval with these models and take the majority vote of the system. 

In addition to adopting different prompts for our models, we also apply the concept of verbalizers~\cite{schick-schutze-2021-exploiting}. More specifically, we use verbalizers to map natural sounding output to the more limited slot-values in the ontology.  For example, given the prompt `Whether the hotel offers wifi', we consider both `True' (or `False') and `Yes' (or 'No') to be the same answer.

\section{Input Example}
(See next page.)

\begin{table*}[h]
\small
\centering
\begin{tabular}{l|p{12cm}}
\toprule
Exemplar 0 (Truncated) &<pad> options available. Would you like to narrow it down by departure time or arrival time? <customer> I'd like to leave after 21:45, if possible. I won't need to book. I'll just need the arrival time, please? <sep> departure of the train is cambridge</s>\\
\midrule
Exemplar 1&taxi destination kambar, taxi departure lovell lodge <agent> when would you like to arrive? <customer> It doesn't matter. I just want to leave there after 10:45 <sep> destination of the taxi is kambar</s>\\
\midrule
Exemplar 2&taxi destination riverboat georgina, taxi departure archway house, hotel area north, hotel day thursday, hotel stay 5, hotel people 3, hotel stars 4, attraction name cambridge punter, attraction type boat <agent> what time would you like to leave or arrive by? <customer> I'd like to leave the hotel by 3:15 please. <sep> stars of the hotel is 4</s>\\
\midrule
Exemplar 3&train day saturday, train destination cambridge, train departure ely <agent> sure, do you know what time you want to arrive? <customer> I want to arrive by 11:30. <sep> departure of the train is ely</s>\\
\midrule
Exemplar 4&restaurant area centre, restaurant people 8, restaurant day thursday, restaurant time 14:00, restaurant food chinese, restaurant price range cheap, taxi destination charlie chan, taxi departure museum of classical archaeology, attraction name museum of classical archaeology <agent> When would you like the leave and arrive by? <customer> I don't mind what time we leave, but I need to arrive at the restaurant by 14:00. <sep> departure of the taxi is museum of classical archaeology</s>\\
\midrule
Exemplar 5&restaurant area south, restaurant food asian oriental, restaurant name any, restaurant price range any, train arrive by none, train day wednesday, train destination cambridge, train departure london kings cross, train leave at none, attraction area east <agent> what time were you wanting to leave by or arrive by? <customer> I want to arrive by 12:15. <sep> arrive by of the train is 12:15</s>\\
\midrule
Prev State & taxi destination pizza hut fen ditton \\
\midrule
Dialog Context & <agent> What time do you want to leave and what time do you want to arrive by? <customer> I want to leave after 17:15. \\
\midrule
Prompt & leave at of the taxi is</s>\\
\midrule
Label&after 17:15\\
\bottomrule
\end{tabular}
\caption{A practical example used during inference which uses our fine-tuned sentence embedder for exemplar retrieval. To be easy to read, we separate each component, including exemplars, query sequence and prompt. Each exemplar contains previous states, dialog context, prompt and label, which corresponds to Sec.~\ref{sec:input}. The 0-th exemplar is truncated so that the entire sequence length can fit into the model.}
\label{table:eg}
\end{table*}